\documentclass[letterpaper, 10 pt, conference]{ieeeconf}  

\IEEEoverridecommandlockouts                              

\usepackage{graphics} 
\usepackage{epsfig} 
\usepackage{amsmath} 
\usepackage{amssymb}  

\usepackage{multirow}
\usepackage{graphicx}
\makeatletter
\providecommand*\caption@documentclass{standard}
\makeatother
\usepackage{cite}
\usepackage{caption}
\usepackage{hyperref} 
\usepackage{color}
\usepackage{algorithm}
\usepackage{algpseudocode}
\usepackage{booktabs}
\usepackage{tabularx}
\usepackage{array}
\usepackage{ragged2e}
\newcolumntype{L}[1]{>{\RaggedRight\arraybackslash}p{#1}}
\newcolumntype{C}[1]{>{\centering\arraybackslash}p{#1}}
\newcolumntype{Y}{>{\RaggedRight\arraybackslash}X}

\usepackage{pifont}

\usepackage[normalem]{ulem}
\useunder{\uline}{\ul}{}

\title{\LARGE \bf
Monocular Vision Based Control Framework for Grasping
}

\author{Shail Jadav$^{1}$ and Dongheui Lee$^{1,2}$
\thanks{$^{1}$Shail Jadav and Dongheui Lee are with Autonomous Systems, Technische Universität Wien (TU Wien), Vienna, Austria (e-mail: \texttt{shail.jadav@tuwien.ac.at, dongheui.lee@tuwien.ac.at}).}%
\thanks{$^{2}$Dongheui Lee is also with the Institute of Robotics and Mechatronics (DLR), German Aerospace Center, Wessling, Germany.}\thanks{This work was supported by the Vienna Science and Technology Fund (WWTF) under the project SafeDiffusion (ICT25068) and  by the European Union project INVERSE (No. 101136067).}
}

\begin{document}

\maketitle
\thispagestyle{empty}
\pagestyle{empty}



\begin{abstract}
Grasping in unstructured environments requires handling objects with widely different mechanical properties, from soft and deformable items to rigid everyday objects. Most existing approaches address these categories separately and often rely on tactile sensing, object-specific models, or specialized grippers. In this paper, we present a unified monocular vision-based grasping framework that targets both soft and rigid objects within a single control pipeline, using only RGB input and a position-controlled gripper. The proposed system combines open-vocabulary object detection, image segmentation, boundary-aware point assignment, real-time point tracking, and monocular depth estimation to recover object motion and geometry from visual observations. A key component of the framework is a language-based stiffness estimation model that infers an object's expected compliance from its semantic description and provides an object-level prior for selecting the grasping strategy before contact. For deformable objects, grasp adaptation is governed by a Procrustes-based dissimilarity measure computed from tracked keypoints, which acts as a visual proxy for deformation. For rigid objects, the gripper width is regulated through the scaling of tracked point distances. We validate the proposed method in real-world pick-and-place experiments on a Franka Emika Research 3 arm using objects with substantially different mechanical properties, including lettuce, fresh mozzarella cheese, croissants, paper towels, and hard plastic bottles. Results demonstrate that the framework achieves stable grasping across both soft and rigid objects using visual feedback alone, highlighting a practical, sensor-efficient, and generalizable approach for food handling and household manipulation.
\end{abstract}
\section{INTRODUCTION}
As robots increasingly integrate into our daily lives, their ability to manipulate various objects, particularly deformable items, becomes crucial \cite{rt2}. These objects include a wide range of items, from food products to flexible materials like cloths and rigid objects. The successful grasping of deformable objects offers significant potential for many industries, especially in food processing and home automation \cite{sanchez2018robotic}. However, the manipulation of deformable objects presents a unique challenge due to their high degrees of freedom and variable material properties \cite{sanchez2018robotic}. Traditional force-based grasping methods are often inadequate for manipulating deformable objects, as they do not consider object deformation \cite{nguyen1988constructing,sanchez2018robotic}. Consequently, advanced sensing and control strategies are necessary to overcome these challenges.\begin{figure}[!t]
    \centering
    \includegraphics[trim=0 00 400 0,width=\linewidth]{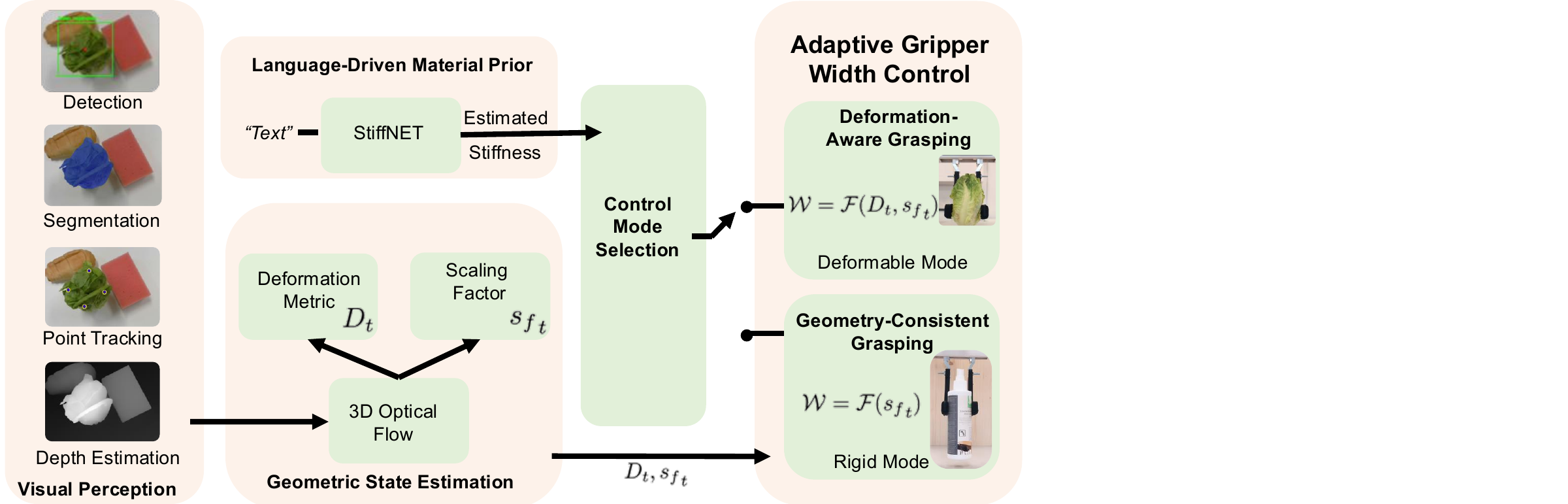}
    \caption{\textbf{Overview of the proposed framework.} A unified monocular vision-based grasping system enables a standard position controlled gripper to handle both compliant and rigid objects using RGB input alone. By combining semantic priors from language with visual feedback during manipulation, the framework selects an appropriate grasping behavior before contact and adapts online to maintain stable grasps across diverse everyday objects.}\vspace{-0.5cm}
    \label{fig:overview}
\end{figure}
One promising advancement is the integration of vision-based tactile sensors in grasping systems \cite{han2024learning,dong2019maintaining,donlon2018gelslim,yuan2017gelsight,fv1,yamaguchi2017implementing}. These sensors operate by using cameras to monitor deformations in elastomeric surfaces, allowing the detection of both normal and shear forces during grasping. While vision-based tactile sensors offer significant advantages in enhancing grasping precision, they also present limitations. Prior studies have shown that objects softer than the sensor elastomer produce weaker and less distinctive tactile signatures, as they undergo greater deformation at the contact interface than the elastomer itself \cite{yueal2016estimating}. Another major drawback is the degradation of elastomers over time with continued use, which leads to reduced sensitivity and performance variability. Furthermore, the cost of these sensors is substantial, including both the initial cost of the system and the ongoing expense of replacing worn elastomers \cite{lambeta2020digit}.

\begin{table*}[!t]
\centering
\footnotesize
\setlength{\tabcolsep}{4pt}
\setlength{\extrarowheight}{0.1cm}
\caption{Comparison with representative prior approaches in terms of their core enabling assumption, use of semantic priors, and demonstrated object regime. Prior work typically derives robustness from tactile instrumentation, explicit mechanics models, or specialized end-effector design. In contrast, the proposed framework operates with monocular RGB and a standard position-controlled gripper while addressing both deformable and rigid objects.}
\label{tab:validated_related_work}

\begin{tabularx}{\textwidth}{@{}L{4.0cm} L{1.8cm} L{5.0cm} C{1.1cm} Y@{}}
\toprule
\textbf{Approach family} &
\textbf{Representative works} &
\textbf{Core enabling assumption} &
\textbf{Semantic prior} &
\textbf{Demonstrated scope} \\
\midrule

Vision--tactile deformable grasping
& \cite{han2024learning}
& Vision-based tactile sensing integrated with a position-controlled gripper
& No
& Deformable-object grasping \\

Vision-based tactile sensing and slip monitoring
& \cite{dong2019maintaining,donlon2018gelslim,yuan2017gelsight,fv1,yamaguchi2017implementing}
& Dedicated tactile fingertips or tactile sensing platforms
& No
& Contact-rich stabilization and tactile behaviors \\

Mechanics-based deformable-object control
& \cite{ficuciello2018fem,fonkoua2024deformation}
& Explicit deformation modeling, often combined with richer 3D perception
& No
& Deformable and soft-object manipulation \\

Semantic adaptive grasping
& \cite{xie_deligrasp_24}
& Force-controllable gripper with gripper-side force/depth sensing
& Yes
& Adaptive grasping from delicate to rigid objects \\

Specialized compliant end-effectors
& \cite{he2020soft,wang20163d,wang2017shape}
& Soft, compliant, or variable-stiffness end-effector design
& No
& Primarily delicate and deformable-object handling \\

\midrule
\textbf{Proposed framework}
& \textbf{This work}
& \textbf{Monocular RGB perception with a standard position-controlled gripper}
& \textbf{Yes}
& \textbf{Unified grasping of deformable and rigid objects} \\

\bottomrule
\end{tabularx}
\setlength{\extrarowheight}{0pt}
\vspace{-0.35cm}
\end{table*}



Another approach to deformable object manipulation involves mechanics based object modelling, such as those utilizing finite element analysis (FEA) \cite{bonet1997nonlinear}. FEA-based methods are commonly used to model strain and control the forces required for manipulating deformable objects \cite{ficuciello2018fem,fonkoua2024deformation,yin2021modeling}. Recent work has also explored using large language models (LLM) to estimate object interaction properties (e.g., mass, friction, and an effective stiffness/compliance proxy) from semantic descriptions to guide grasping policies \cite{xie_deligrasp_24}. These methods can be highly effective when the material properties and environmental dynamics are accurately modelled. However, in real-world applications, obtaining precise material properties is challenging. Many deformable objects exhibit time-varying and nonlinear behaviors, making it difficult to maintain accurate models over time and across varying conditions. For example, material properties such as stiffness in food products change over time due to factors like moisture content or age, complicating the modelling process \cite{goranova2015sensory}. 

Recent advancements in gripper technology have focused on the development of specialized structures for grasping deformable objects \cite{he2020soft}. In particular, soft grippers have shown significant promise in handling delicate items such as food products \cite{wang20163d}. These grippers are typically actuated either by pneumatic pressure or tendons, and some utilize shape memory alloys for actuation \cite{wang2017shape}. Due to their compliance, soft grippers are well-suited for tasks requiring gentle handling, as they do not exert large forces, which is ideal for fragile objects. However, this very compliance poses a limitation when it comes to the generalization of these grippers. Specifically, they struggle to generate the necessary forces required for manipulating rigid objects effectively, thereby limiting their versatility in more demanding applications\cite{trinh2024novel}.

As advancements in robotic grasping technologies continue to progress, significant inspiration can be drawn from the way humans manipulate objects. Humans skillfully use the same hands to handle a wide variety of objects, both deformable and rigid, by seamlessly integrating visual and tactile feedback \cite{johansson2009coding,camponogara2019grasping,sobinov2021neural}. While tactile feedback is crucial for fine manipulation and adjusting grip forces, especially during rapid movements where visual feedback may lag, humans can still perform grasping tasks relying solely on visual feedback, albeit with reduced precision and increased effort compared to when both sensory modalities are utilized \cite{augurelle2003importance,nowak2003selective}. Beyond perception, humans also exploit semantic knowledge: object and material words can provide a prior over expected mechanical properties (e.g., soft vs.\ stiff), with material names alone eliciting a structured softness representation comparable to visually and haptically derived spaces \cite{cavdan2023assessing}.
Inspired by this human adaptability, we aim to develop a framework that relies exclusively on visual feedback for robotic grasping tasks. This approach seeks to simplify sensory requirements and enhance the applicability of robotic systems in environments where tactile sensing is impractical or unavailable. 

With advancements in grasping technologies, there have also been significant breakthroughs in the field of computer vision, which can greatly enhance object-grasping capabilities. Recent progress in object detection and image segmentation is exemplified by the release of SAM 2, a transformer-based foundational model for image segmentation, which excels at segmenting objects of interest in cluttered environments \cite{sam2}. For grasping deformable objects, real-time state tracking can be achieved using point-tracking algorithms like TAPIR, developed by Google DeepMind, which tracks individual query points \cite{tapir}. On the other hand, Meta's CoTracker, designed for tracking multiple points simultaneously, is more suited for rigid objects \cite{karaev2023cotracker}. Moreover, recent developments in depth estimation models enable the inference of relative depth from RGB images \cite{depth_anything_v2}. By combining 2D point tracking with relative depth information, one can track query points on deformable objects in real-time and in 3D space, offering valuable insights into object deformation during grasping. This approach could be utilized to dynamically adjust the gripper's width for improved control during grasping tasks.
As a result, this paper presents a monocular vision-based framework for grasping both deformable and rigid objects without tactile sensing, specialized grippers, or explicit mechanics-based object models, as illustrated in Fig.~\ref{fig:overview} and Table \ref{tab:validated_related_work}. The core technical contributions of this work are as follows:
\begin{itemize}
    \item We present a unified monocular vision-based grasping framework that enables a standard position controlled gripper to handle both compliant and rigid objects using RGB input alone.
    \item We show that semantic knowledge from language can provide useful object-level priors about compliance before contact, allowing the system to choose an appropriate grasping behavior without tactile sensing or explicit mechanics models.
    \item We demonstrate that visual feedback alone can support adaptive and stable grasping across a broad range of everyday objects, highlighting a practical and sensor-efficient approach for food handling and household manipulation.
\end{itemize}

\begin{figure}[!t]
    \centering
    \includegraphics[trim=00 00 00 200,width=0.65\linewidth]{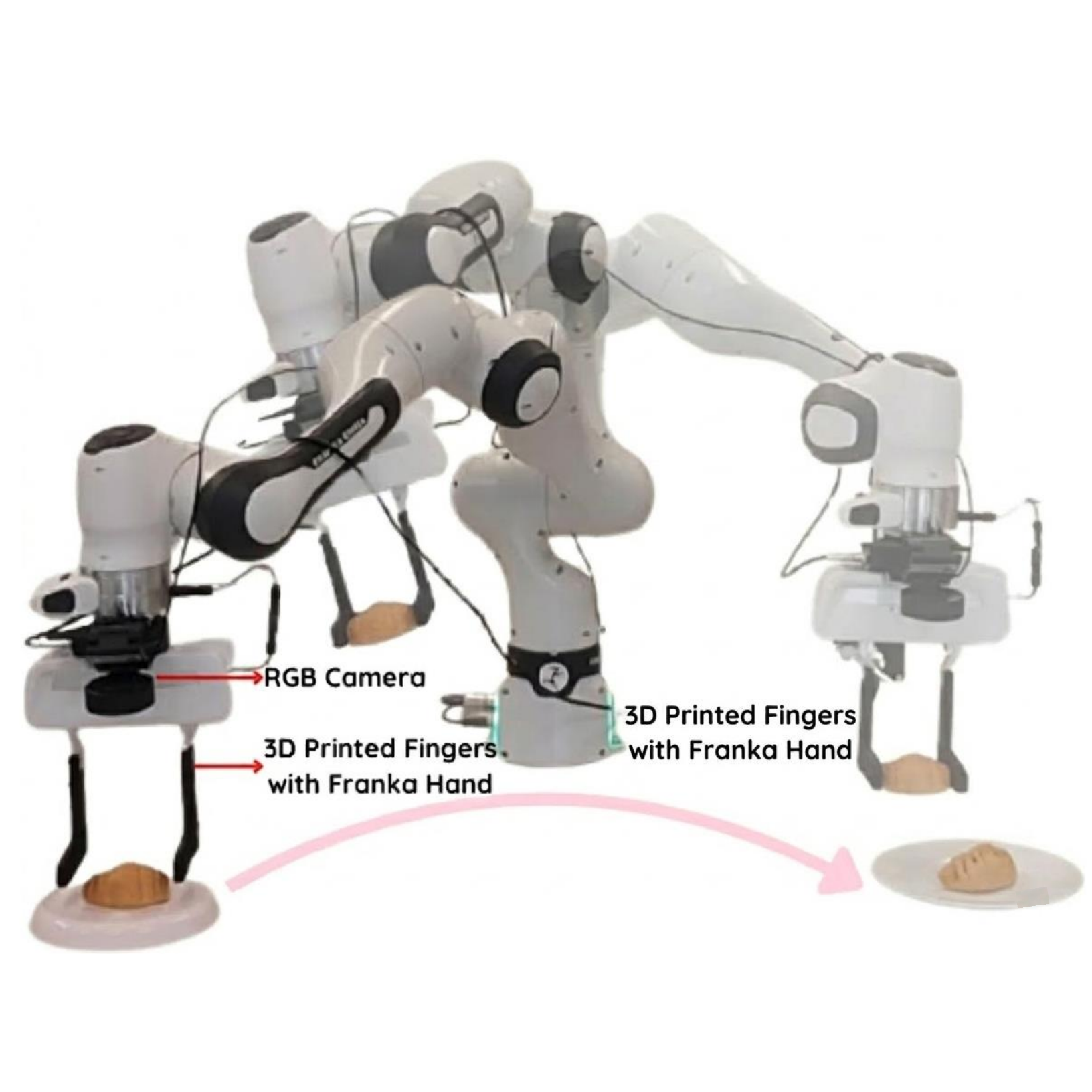}
    \vspace{-0.7cm}
    \caption{\textbf{Experimental setup for grasping \& placing experiment}:Involving a generic USB camera, 3D-printed fingers, and a Franka arm with gripper.}
    \label{fig:setup}
    \vspace{-0.7cm}
\end{figure}

Finally, we validate the proposed system through a series of experiments with real-world objects, showcasing its potential in real-world scenarios. The framework is implemented using a position-controlled Franka Emika Hand (gripper), allowing for precise grasping, as shown in Fig. \ref{fig:setup}. We evaluate the performance of our method on objects that are challenging to manipulate, such as fresh mozzarella cheese, lettuce, croissants, and paper towels, as well as rigid objects like hard plastic bottles, as shown in Fig \ref{fig:mul_obj}. These objects present unique challenges for robotic grasping due to their diverse physical properties. Lettuce and croissants exhibit nonhomogeneous and anisotropic material properties, leading to unpredictable mechanical responses during manipulation. Fresh mozzarella cheese is slippery, introducing slippage issues that complicate stable grasping. Paper towels and rigid plastic bottles serve as examples of everyday items with differing rigidity and surface textures, further testing the adaptability and robustness of our grasping method. Our framework successfully completed pick-and-place operations with these objects, demonstrating its potential for food handling and home automation applications.

\section{METHODS}
In this section, we present our framework for grasping both deformable and rigid objects. The object detection algorithm receives a prompt of the object of interest and camera feed as input, identifying the object's location within the frame. This location is then passed to an image segmentation network, which isolates the object of interest from the background. The resulting segmented mask is forwarded to the grid point assignment algorithm, where query points for tracking are selected. These tracking query points are transmitted to a tracking network, providing real-time 2D coordinates within the image. Depth information is subsequently acquired using a depth estimation model, and with both 2D coordinates and depth, a full 3D representation of the object is constructed. Procrustes analysis is then performed by comparing the current tracking points with the points from the initial conditions to assess dissimilarity, which correlates with the force exerted by the gripper on deformable objects. This dissimilarity arises from the deformation of key points within the object. The gripper controller uses this dissimilarity to adapt the gripper's width, ensuring a stable grasp. For rigid objects, the gripper width is directly controlled based on the scaling factor derived from the ratio of the pairwise distance between the tracked points to their pairwise distance at the initial condition. The system utilizes monocular vision and real-time control of the Franka Hand. An overview of the proposed framework is depicted in Fig. \ref{fig:overview}.\vspace{-0.1cm} 
\subsection{Object Detection \& Segmentation}
Let $\mathbf{I} \in \mathbb{R_{+}}^{3 \times H \times W}$ represent the input RGB image frame, and let $\mathbf{p}$ denote the object of interest $\mathbf{O}$. The image frame $\mathbf{I}$, together with the textual prompt $\mathbf{p}$ specifying the object of interest, is passed to the object detection algorithm. Our framework employs the YOLOv8x-worldv2 model for real-time open-vocabulary object detection \cite{yolov8}. The output of the detection process provides the coordinates of \(\mathbf{O}\) as \vspace{-0.1cm}
\begin{align}
    \mathbf{o}_{c} = (x_2 - c, \frac{y_1 + y_2}{2}),
    \label{eq:oc}\vspace{-0.3cm}
\end{align}
where \(x_2, y_1,\text{\& } y_2\) are the bounding box coordinates for the object $\mathbf{O}$, and \(c \in \mathbb{Z}_{+}\) is a user-defined constant scalar value, allowing the selection of segments near the edge of the bounding box. The coordinates \(\mathbf{o}_{c}\in\mathbb{Z}^{2}_{+}\) are passed to the Meta SAM2 (Segment Anything Model \cite{sam2}) for image segmentation, which generates the mask \(\mathbf{M} \in \{0,1\}^{H \times W}\) corresponding to \(\mathbf{O}\). The mask \(\mathbf{M}\) is then provided to the point assignment algorithm to create query points for the tracking point network. Here we employ Tracking Any Point with per-frame Initialization and Temporal Refinement (TAPIR) model for point tracking \cite{tapir}. 
Because point tracking degrades on low-texture objects, query-point assignment is formulated as a boundary-weighted clustering problem that selects evenly distributed, maximally separated points near the mask contour to improve tracking robustness.
\begin{figure*}[!htpb]
    \centering
    \includegraphics[width=0.9\linewidth]{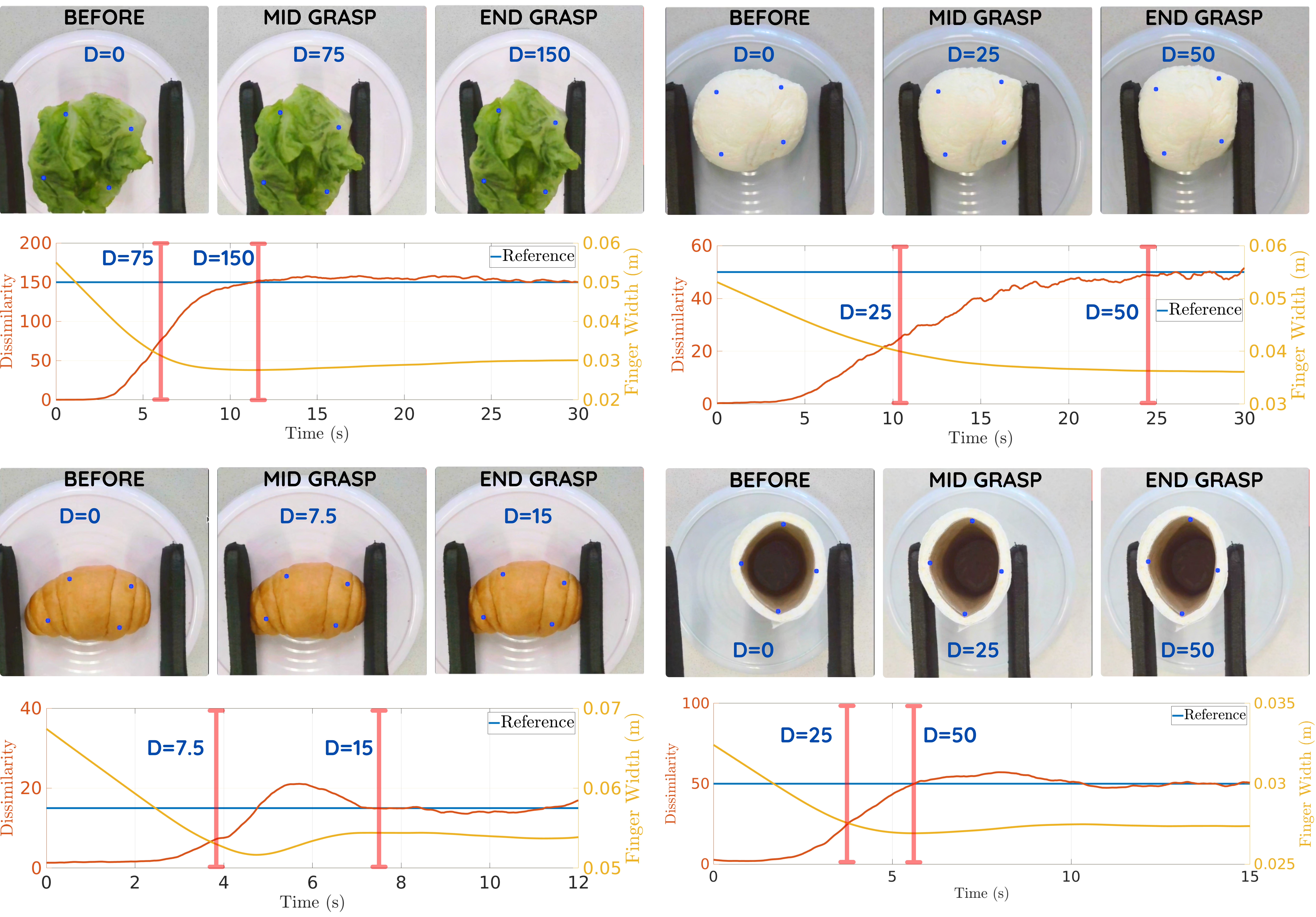}
    \caption{\textbf{Grasping multiple objects}: We demonstrate successful grasping during a pick-and-place operation for deformable objects, including lettuce, mozzarella cheese, croissant bread, and a paper roll. The proposed controller adjusts the gripper finger width to track the desired dissimilarity and maintain a stable grasp throughout the manipulation. D indicates the dissimilarity of the tracking points compared to the initial condition, and red lines show the instances of particular dissimilarity values.}
    \label{fig:mul_obj}
    \vspace{-0.75cm}
\end{figure*}
\vspace{-0.2cm}\subsection{Point Assignment}
We begin by extracting the contours from the binary mask \(\mathbf{M}\), which defines the region of interest. The contours represent the boundaries between these distinct regions in the mask. From the detected contours, we select the largest contour, denoted \(\mathcal{C}\), based on the area enclosed by the contour. Next, we gather all the points within the mask where the pixel value is positive. These points form the set \(\mathcal{P}\), which is defined as:\vspace{-0.2cm}
\begin{align}
    \mathcal{P} = \{(x, y) \mid \mathbf{M}(x, y) > 0 \}.
\end{align}
For each point \( \mathbf{p} = (x, y) \in \mathcal{P} \), we compute its distance to the nearest contour point \( \mathbf{c} \in \mathcal{C} \). This distance is defined as\vspace{-0.2cm}
\begin{align}
    d(\mathbf{p}) = \min_{\mathbf{c} \in \mathcal{C}} \|\mathbf{p} - \mathbf{c}\|.
\end{align}
We assign a weight to each point \( \mathbf{p} \in \mathcal{P} \) inversely proportional to its distance from the contour as
\begin{align}
w(\mathbf{p}) = \frac{1}{1 + d(\mathbf{p})}.
\end{align}

This weighting scheme emphasizes points closer to the contour, assigning them larger weights. We then apply a weighted \( k \)-means clustering algorithm to partition the set of points \( \mathcal{P} \) into \( k \) clusters, where \( k \) is the desired number of grid points selected by the user. The \( k \)-means clustering algorithm seeks to minimize the within-cluster variance by optimizing the placement of cluster centers \( \mathbf{G} = \{\mathbf{g}_1, \mathbf{g}_2, \ldots, \mathbf{g}_k\} \), which will serve as the final grid points. The objective function to be minimized is\vspace{-0.2cm}
\begin{align}
   \mathcal{J}(\mathbf{G}) = \sum_{i=1}^{k} \sum_{\mathbf{p}_j \in \mathcal{K}_i} w(\mathbf{p}_j) \|\mathbf{p}_j - \mathbf{g}_i\|^2,
\end{align}
where \( \mathbf{p}_j \in \mathcal{P} \) represents a point in the mask, \( \mathcal{K}_i \) is the set of points assigned to the \( i \)-th cluster, \( \mathbf{g}_i \in \mathbb{R}^2 \) is the center of the \( i \)-th cluster, and \( w(\mathbf{p}_j) \) is the weight of point \( \mathbf{p}_j \). The cluster centers \( \mathbf{g}_i \) are updated iteratively using the weighted mean of the points assigned to each cluster \vspace{-0.1cm}
\begin{align}
\mathbf{g}_i = \frac{\sum_{\mathbf{p}_j \in \mathcal{K}_i} w(\mathbf{p}_j) \mathbf{p}_j}{\sum_{\mathbf{p}_j \in \mathcal{K}_i} w(\mathbf{p}_j)}.
\end{align}
This process repeats until convergence, ensuring that the cluster centers minimize the weighted variance of points in each cluster. As a result, the grid points \( \mathbf{G} \) are distributed across the mask with a denser concentration near the contour due to the weighting scheme. After clustering, we verify that each grid point \( \mathbf{g}_i = (x_i, y_i) \in \mathbf{G} \) lies within the mask \( \mathbf{M} \). Specifically, for each \( \mathbf{g}_i \), we check \(\mathbf{M}(x_i, y_i) > 0\). If any grid points lie outside the mask, they are discarded. If the number of valid grid points is less than the desired number \( k \), we iteratively refine the grid by adding new points. To do this, we identify the pair of points \( (\mathbf{g}_i, \mathbf{g}_j) \in \mathbf{G} \) that are farthest apart \(
(\mathbf{g}_i, \mathbf{g}_j) = \arg\max_{i,j} \|\mathbf{g}_i - \mathbf{g}_j\|\). The midpoint between \( \mathbf{g}_i \) and \( \mathbf{g}_j \) is computed as \(
\mathbf{g}_{\text{new}} = \frac{\mathbf{g}_i + \mathbf{g}_j}{2}.
\) If \( \mathbf{g}_{\text{new}} \in \mathbf{M} \), it is added to the set \( \mathbf{G} \); otherwise, the closest valid point within \( \mathcal{P} \) is selected. The final set of grid points \( \mathbf{G} \) is distributed within the mask, with emphasis on regions closer to the contour.\vspace{-0.15cm}
\subsection{Point Tracking and Depth Estimation}
The grid points \(\mathbf{G}\), generated during the point assignment process, serve as input query points to the TAPIR model, which is optimized for efficiently tracking arbitrary points in video sequences. TAPIR operates in two stages. In the matching stage, it independently identifies candidate matches for each query point across frames. Then, in the refinement stage, it updates the point trajectories and query features by utilizing local correlations, improving both tracking accuracy and consistency over time. The output of TAPIR is the set of tracked points \(\mathbf{G}^{t}\), where \(t\) represents time. Once real-time 2D tracking points are obtained from TAPIR, we apply the Depth Anything model for monocular depth estimation, converting the 2D points into 3D tracking points in space \cite{depth_anything_v2}. Both TAPIR and Depth Anything operate in real-time, enabling dynamic point tracking even in complex environments. Additionally, depth can be inferred using a scaling factor, assuming the object remains within the same plane of reference and does not rotate. This scaling factor assists in depth estimation, while the Depth Anything model provides relative depth estimates, which enhance the determination of object orientation in 3D space.\vspace{-0.15cm}
\subsection{Procrustes Analysis}
Procrustes analysis is used to compare the shapes of two data matrices by optimally transforming one to match the other. The objective is to minimize the dissimilarity between the two matrices. This transformation involves applying translations, rotations, reflections, and scaling to achieve the best fit. In this context, we calculate the dissimilarity between the initial grid points \( \mathbf{G}^{0} \) and the current grid points \( \mathbf{G}^{t} \). The \(\mathbf{G}^{0} \) is the output of the point assignment algorithm and \(\mathbf{G}^{t}\) is the output of TAPIR with Depth Anything model, only TAPIR used if orientation is not required. First, both grid points are centered by subtracting their mean vectors, ensuring that their centroids coincide with the origin \vspace{-0.15cm}
\begin{align}
      \mathbf{G}^0_c = \mathbf{G}^0 - \mathbf{1}_n \bar{\mathbf{G}^0}, \quad \mathbf{G}^t_c = \mathbf{G}^t - \mathbf{1}_n \bar{\mathbf{G}^t}, 
\end{align}
where \( \bar{\mathbf{G}^0} \) and \( \bar{\mathbf{G}^t} \) are the mean vectors of the respective grid points, and \( \mathbf{1}_n \) is an \( n \)-dimensional vector of ones. Further, we normalize the  grid points as \vspace{-0.2cm}
\begin{align}
    \mathbf{G}^0_n = \frac{\mathbf{G}^0_c}{\|\mathbf{G}^0_c\|}, \quad \mathbf{G}^t_n = \frac{\mathbf{G}^t_c}{\|\mathbf{G}^t_c\|}.
\end{align}
After the normalization, we find the optimal rotation and scaling factor for the minimal dissimilarity. We perform singular value decomposition to find optimal rotation matrix and scaling factor as \vspace{-0.3cm}
\begin{align}
    \mathbf{U},\mathbf{S},\mathbf{V}^\top=\text{SVD}({\mathbf{G}^0_n}^\top \mathbf{G}^t_n).
\end{align} The optimal rotation matrix \( \mathbf{R} \) and the scaling \( s \) are given by \vspace{-0.5cm}
\begin{align}
   \mathbf{R} = \mathbf{U} \mathbf{V}^\top, \quad s = \sum \mathbf{S}. 
\end{align}
Finally, the dissimilarity \( D_t \) between the grid points is computed as \vspace{-0.7cm}
 \begin{align}
    D_t = \Gamma\sum_{i=1}^{k} \|{\mathbf{G}^0_n}^{(i)} - s \mathbf{R}{\mathbf{G}^t_n}^{(i)} \|^2 , 
 \end{align}
 where \(\Gamma\in\mathbb{R}_{+}\) is the sensitivity gain. The dissimilarity \( D_t \) provides a quantitative assessment of the dissimilarity between the two gridpoints, where smaller values of \( D_t \) indicate greater similarity. We further define scaling factor as \vspace{-0.55cm}
\begin{align}
   {s_f}_t = \frac{\sqrt{\sum_{i=1}^{k} \sum_{j=1}^{k} \|\mathbf{g}^t_i - \mathbf{g}^t_j\|^2}}{\sqrt{\sum_{i=1}^{k} \sum_{j=1}^{k} \|\mathbf{g}^0_i - \mathbf{g}^0_j\|^2}},
\end{align}
where \(g^t\in\mathbf{G}^{t}\) and \(g^0\in\mathbf{G}^{0}\) are the individual grid points. 

\begin{figure}[!b]
    \centering
    \includegraphics[width=0.9\linewidth]{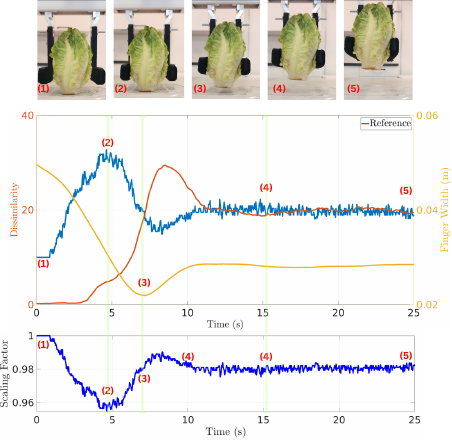}
    \caption{\textbf{Grasping lettuce}: During the robot's upward movement, a decrease in the scaling factor is observed when the lettuce is not grasped. This reduction increases the reference dissimilarity, leading to a decrease in the gripper's finger width. Consequently, the gripper gradually closes until the lettuce is grasped; after grasping, all parameters stabilize. Red numbers and green lines in the figures indicate specific instances in the experiments.}
    \label{fig:veggrasp}
\end{figure}

\subsection{Language-Based Stiffness Estimation (StiffNET)  }
\label{sec:lang_stiffness}

We estimate object stiffness directly from language by learning a scalar mapping from an object's text description to its physical stiffness. The key idea is to embed each object name into a semantic feature space and then learn a function that places objects on a one-dimensional \emph{log-stiffness axis}. Training combines two complementary supervision sources: 1) pairwise hardness comparisons, which provide relative ordering, and 2) sparse ground-truth stiffness measurements, which anchor the absolute scale.

\paragraph{Text representation.}
Let \(x \in \mathcal{X}\) denote the text associated with an object, such as its category name or short semantic descriptor. A pretrained text encoder \(\mathcal{E}(\cdot)\) maps \(x\) to a normalized embedding
\begin{equation}
    \mathbf{z}(x) = \mathcal{E}(x) \in \mathbb{R}^{d},
    \qquad \|\mathbf{z}(x)\|_2 = 1.
\end{equation}
In our implementation, \(\mathcal{E}\) is a frozen CLIP text encoder. Thus, the semantic representation is fixed during training, and only the downstream stiffness predictor is learned.

\paragraph{Stiffness network.}
The main trainable model is a neural network \(f_{\theta} : \mathbb{R}^{d} \rightarrow \mathbb{R}\) that predicts the \emph{log-stiffness} of an object:
\begin{equation}
    \hat{\ell}(x) = f_{\theta}(\mathbf{z}(x)) + b_{\theta},
\end{equation}
where \(b_{\theta} \in \mathbb{R}\) is a learned scalar bias. The predicted stiffness is then recovered as
\begin{equation}
    \hat{k}(x) = \exp\!\big(\hat{\ell}(x)\big).
\end{equation}
Predicting \(\log\)-stiffness rather than stiffness directly improves numerical conditioning and is more suitable when stiffness spans multiple orders of magnitude.

\paragraph{Pairwise comparison supervision.}
We first consider a dataset of pairwise hardness comparisons
\begin{equation}
    \mathcal{D}_{P}
    =
    \left\{
    \big(x^{(1)}_{p}, x^{(2)}_{p}, y_{p}, c_{p}\big)
    \right\}_{p=1}^{N_{P}},
\end{equation}
where \(x^{(1)}_{p}\) and \(x^{(2)}_{p}\) are the two objects in comparison \(p\), \(y_{p}\in\{+1,-1\}\) indicates which object is harder, and \(c_{p}\in[0,1]\) is the annotation confidence. Specifically, \(y_{p}=+1\) means \(x^{(1)}_{p}\) is harder than \(x^{(2)}_{p}\), and \(y_{p}=-1\) means the opposite. We generate this dataset by prompting LLM (Gemma 3) to compare pairs of daily life objects and provide confidence scores for its comparisons. For example,  "\textit{Which is harder, a banana or a metal bolt}?" and ask it to provide a confidence score for its answer. This approach allows us to leverage the LLM's extensive world knowledge to generate a rich set of pairwise comparisons without requiring manual annotation. For each pair, the stiffness network predicts
\begin{equation}
    \hat{\ell}^{(1)}_{p} = \hat{\ell}\!\left(x^{(1)}_{p}\right),
    \qquad
    \hat{\ell}^{(2)}_{p} = \hat{\ell}\!\left(x^{(2)}_{p}\right),
\end{equation}
and the corresponding predicted log-stiffness difference is
\begin{equation}
    \Delta \hat{\ell}_{p}
    =
    \hat{\ell}^{(1)}_{p} - \hat{\ell}^{(2)}_{p}.
\end{equation}
If \(y_p=+1\), the desired outcome is \(\Delta \hat{\ell}_p > 0\); if \(y_p=-1\), the desired outcome is \(\Delta \hat{\ell}_p < 0\). Thus, pairwise supervision teaches the network the \emph{relative ordering} of objects along the stiffness axis.

\paragraph{Auxiliary margin network.}
To allow different comparison pairs to have different separation requirements, we introduce an auxiliary margin network \(h_{\phi}\). This network is used only during training; it is not needed at inference time. For pair \(p\), we define an auxiliary GT-derived feature
\begin{equation}
    \Delta \ell^{\star}_{p}
    =
    \begin{cases}
        \left|
        \log k^{\star}\!\left(x^{(1)}_{p}\right)
        -
        \log k^{\star}\!\left(x^{(2)}_{p}\right)
        \right|,
        & \text{if ground-truth,}
        \\[4pt]
        0, & \text{otherwise,}
    \end{cases}
\end{equation}
where \(k^{\star}(\cdot)\) denotes measured stiffness. The margin network receives the concatenated feature vector
\begin{equation}
    \mathbf{u}_{p}
    =
    \left[
    \mathbf{z}\!\left(x^{(1)}_{p}\right);
    \mathbf{z}\!\left(x^{(2)}_{p}\right);
    \Delta \ell^{\star}_{p}
    \right]
    \in \mathbb{R}^{2d+1}.
\end{equation}
and predicts a pair-specific margin score
\begin{equation}
    \hat{m}_{p} = h_{\phi}(\mathbf{u}_{p}),
\end{equation} where \(h_{\phi}\) is neural network. This design allows the model to adapt the ranking margin to the semantic content of the pair and, when available, to the magnitude of known stiffness differences. The pairwise supervision is enforced through a hinge loss in log-space:
\begin{equation}
    \mathcal{L}_{\mathrm{rank}}
    =
    \frac{1}{|\mathcal{B}_{P}|}
    \sum_{p \in \mathcal{B}_{P}}
    \beta_{p}
    \left[
    \hat{m}_{p}
    -
    y_{p}\Delta \hat{\ell}_{p}
    \right]_{+},
\end{equation}
where \([a]_{+}=\max(a,0)\), \(\mathcal{B}_{P}\) is a minibatch of pairwise samples. This loss is zero whenever the predicted ordering is correct and the signed log-stiffness difference exceeds the required margin. Otherwise, it pushes the stiffness network to move the harder object upward and the softer object downward on the learned log-stiffness axis.

\paragraph{Ground-truth regression supervision.}
Pairwise supervision alone is insufficient to recover an absolute stiffness scale, since it only constrains relative ordering. To anchor the predictions numerically, we use a second dataset of sparse ground-truth stiffness values
\begin{equation}
    \mathcal{D}_{G}
    =
    \left\{
    \big(x_{q}, k^{\star}_{q}\big)
    \right\}_{q=1}^{N_{G}},
\end{equation}
where \(k^{\star}_{q}>0\) is the measured stiffness of object \(x_q\). The corresponding log-space target is
\begin{equation}
    \ell^{\star}_{q} = \log(k^{\star}_{q} + \varepsilon).
\end{equation} We define the regression loss as
\begin{equation}
    \mathcal{L}_{\mathrm{reg}}
    =
    \frac{1}{|\mathcal{B}_{G}|}
    \sum_{q \in \mathcal{B}_{G}}
    \,
    \mathrm{Huber}
    \big(
    \hat{\ell}(x_{q}),
    \ell^{\star}_{q}
    \big),
\end{equation}
where \(\mathcal{B}_{G}\) is a minibatch of GT objects. This term directly pulls the predicted log-stiffness toward measured values and therefore determines the absolute scale of the learned stiffness axis.

\paragraph{Joint training of the stiffness network.}
The stiffness network \(f_{\theta}\) is trained jointly by the pairwise ranking loss and the GT regression loss:
\begin{equation}
    \mathcal{L}
    =
    \lambda_{\mathrm{rank}}\,\mathcal{L}_{\mathrm{rank}}
    +
    \lambda_{\mathrm{reg}}\,\mathcal{L}_{\mathrm{reg}}.
\end{equation}
The critical point is that the stiffness network receives gradients from \emph{both} terms. The ranking loss teaches \emph{relative order}, while the regression loss teaches \emph{absolute scale}. Consequently, the learned predictor does not merely separate hard from soft objects; it also places them at physically meaningful locations in log-stiffness space.

In contrast, the auxiliary margin network \(h_{\phi}\) is updated only through \(\mathcal{L}_{\mathrm{rank}}\). Its purpose is to shape the ranking constraint during training, whereas the final stiffness prediction is entirely produced by \(f_{\theta}\).

\begin{table}[b]
\centering
\vspace{-0.5cm}
\caption{Comparison of Young's modulus estimates (GPa) produced by GPT-5.3 and \textbf{StiffNET}.}
\footnotesize
\setlength{\tabcolsep}{4pt}
\renewcommand{\arraystretch}{1.10}
\begin{tabular}{lccc}
\toprule
\multirow{2}{*}{\textbf{Object}} & \multirow{2}{*}{\textbf{True (GPa)}} & \multirow{2}{*}{\textbf{GPT-5.3}} & \textbf{StiffNET} \\
&  &  & \textbf{Prediction} \\
\midrule
Tofu                      & 0.0001 & 0.0001 & 0.0001002 \\
Wooden Block              & 10     & 10     & 9.092 \\
Steel                     & 200    & 200    & 210.3 \\
PVC (Polyvinyl Chloride)  & 2.5    & 2.5    & 2.568 \\
\midrule
Cucumber                  &  -     & 0.002  & 0.03191 \\
Walnut                    &  -     & 6      & 18.18 \\
Scissors                  &  -     & 200    & 157 \\
Power Bank                &  -     & 3      & 5.451 \\
Plastic Bottle            &  -     & 1.5    & 2.1 \\
\bottomrule
\end{tabular}

\label{tab:youngs_modulus_comparison}
\end{table}

\subsection{Grasp Control}
StiffNET provides an object-specific prior on mechanical behavior directly from language, enabling the controller to select an appropriate grasping strategy before contact and without requiring tactile sensing or explicit material models. Specifically, the stiffness estimate \(\hat{k}\) produced by StiffNET is compared against a threshold \(k_{\mathrm{th}}\): objects with \(\hat{k} < k_{\mathrm{th}}\) are treated as deformable/compliant, whereas objects with \(\hat{k} \geq k_{\mathrm{th}}\) are treated as rigid. Inspired by the way humans use semantic priors during manipulation, we then employ a control strategy that adapts the gripper width using visual feedback. For deformable objects, the dissimilarity of the tracked points is used as a proxy for interaction-induced deformation and, consequently, the applied grasp force. In parallel, the scaling factor captures changes in the relative distance between the object and the camera, which is particularly informative for rigid-object grasping.

The grasp controller is given by\vspace{-0.2cm}
\begin{align}
    \mathcal{W}_{t+1} = \mathcal{W}_{t} - \lambda({D_{ref}}_t - D_t) ,
\end{align}
where \(\mathcal{W}_{t}\) is commanded gripper width at time $t$, \( \lambda \in \mathbb{R}_{+} \) is the adaptation gain, and \vspace{-0.1cm}
\begin{align}
    {D_{ref}}_t = D_{min} + \Omega(1 - {s_f}_t) \label{dmin}
\end{align}
is the reference dissimilarity. Here, \( D_{min} \) represents the constant minimum dissimilarity set by the user or also can be estimated through StiffNET, \( \Omega \in \mathbb{R}_{+} \) is the scaling factor gain, and \( {s_f}_t \) is the scaling factor. Initially, the controller produces a desired dissimilarity based on \( D_{min} \). If the object begins to slip during grasping, the scaling factor \( {s_f}_t \) decreases, causing \( D_{ref} \) to increase. The gripper width is then adjusted proportionally until the slipping stops, resulting in a secure and stable grasp of the deformable object. While this control strategy is effective for deformable objects, it is not suitable for rigid objects since dissimilarity remains minimal due to the lack of deformation. For rigid objects, the control relies on the scaling factor and is expressed as\vspace{-0.1cm}
\begin{align}
     \mathcal{W}_t = ({s_f}_{t} - {s_f}_{min}) \frac{ (\mathcal{W}_{max} - \mathcal{W}_{min})}{ (1 - {s_f}_{min})} + \mathcal{W}_{min},
\end{align}
where \( \mathcal{W}_{max} \) and \( \mathcal{W}_{min} \) are the gripper's maximum and minimum widths, and \( {s_f}_{min} \) is the minimum scaling factor and a user-defined sensitivity parameter to control the gripper width. This strategy allows for the gripper width to be updated until the scaling factor remains variable, ensuring a secure grasp of rigid objects.\vspace{-0.1cm}
\section{EXPERIMENTS}
To validate the performance of the proposed framework, we conducted grasping experiments using various objects. These included deformable objects such as lettuce, mozzarella cheese, croissant bread, and paper towels, as well as a rigid object like a hard plastic bottle. For these experiments, we used a Franka Emika Research 3 robotic arm equipped with a Franka Hand, as shown in Fig \ref{fig:setup}. A RAZER Kiyo-X generic webcam was used for vision input. Additionally, we extended the fingers of the Franka Hand with 3D-printed PLA extensions with foam cushions. The models were deployed on an Nvidia RTX A4000 GPU, and with four grid points (\(k=4\)), we were able to run the algorithm at 30 frames per second.

In the first experiment involving lettuce, we passed the prompt \textit{green vegetables} to the object detection algorithm, which subsequently provided the coordinates for object segmentation. We used $k_{\mathrm{th}}=0.1$ and set the minimum dissimilarity as \( D_{min} = 10 \), after which the robot began the grasping process. During grasping, the robot exhibited upward motion, causing the scaling factor to decrease, as depicted in Fig. \ref{fig:veggrasp}. This reduction in the scaling factor led to an increase in the reference dissimilarity, \( {D_{ref}}_t \), as the scaling factor gain was defined by \( \Omega = 500 \). The adaptation gain for the finger width was set to \( \lambda = 7 \times 10^{-6} \). Other parameters are as follows \(\Gamma=10^4\) and \(c=5\). The scaling factor converges as the robot achieves a stable grasp, allowing it to successfully execute the pick-up operation. It is important to highlight that the finger width presented in the results corresponds to the commanded finger width. Due to the physical structure of the Franka Hand, when the commanded finger width is set to zero, the actual physical finger width remains at 0.02 meters. It is important to note that the measured dissimilarity can vary across trials and across objects (e.g., the lettuce trials in Fig.~\ref{fig:veggrasp} versus the multi-object results in Fig.~\ref{fig:mul_obj}) due to variation in object properties. Therefore, we use the estimated stiffness only for control-mode selection.

\begin{figure}[!t]
    \centering
    \includegraphics[width=0.9\linewidth]{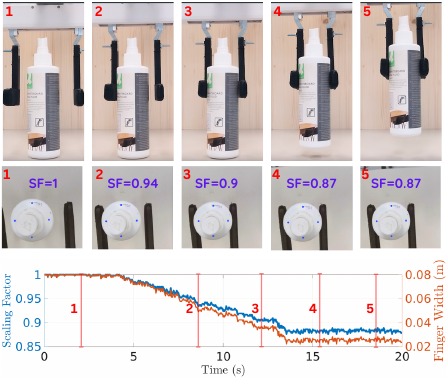}
    \caption{\textbf{Grasping bottle}: As the robot ascends, the distance between the camera and the bottle's tracking points increases, reducing the scaling factor. This leads the gripper controller to command narrower finger widths. Once the bottle is grasped, both the scaling factor and finger width stabilize. The top row of images illustrates the robot's upward motion during the bottle-grasping process, while the bottom row shows the scaling factor (SF) and finger width. Red markings indicate instances in the experiment. }
    \label{fig:botteimg}
    \vspace{-0.8cm}
\end{figure}

In our grasping experiments, the desired dissimilarity for any object is not known a priori. However, by conducting an initial grasping trial with upward movement, we can observe at what point the dissimilarity stabilizes.  We then use this stabilized value as the minimum dissimilarity for subsequent grasping attempts. It is important to note that this approach does not account for dynamic cases, such as scenarios involving high acceleration, which could lead to slippage during grasping. Therefore, to ensure robustness, a safety factor should be introduced when determining the minimum dissimilarity. This safety factor is multiplied with \(D_{min}\) in (\ref{dmin}), which compensates for higher accelerations that may occur during manipulation. For example, during our experiments with lettuce, we applied a safety factor of 7.5. Using this method, we successfully demonstrate pick-and-place manipulation across multiple objects, as shown in Fig. \ref{fig:mul_obj}. In each trial, the system first adjusts to the desired dissimilarity before performing a successful manipulation. The finger width adapts accordingly, ensuring desirable tracking performance and stable grasps.

We further demonstrated grasp manipulation with a rigid object, selecting a hard plastic bottle for the experiment. For the rigid object manipulation, we chose the following parameters: minimum scaling factor \({s_f}_{\text{min}} = 0.85\), maximum gripper width \(\mathcal{W}_{\text{max}} = 0.08 \, \text{m}\), and minimum gripper width \(\mathcal{W}_{\text{min}} = 0.01 \, \text{m}\). The experiment begins with the gripper fully open, after which the robot moves upward. This movement causes a reduction in the scaling factor, \({s_f}_t\), resulting in a decrease in finger width until the gripper successfully grasps the bottle. Once the bottle is grasped, the relative distance between the bottle and the camera remains fixed, leading to the stabilization of both the scaling factor and the finger width, as shown in Fig. \ref{fig:botteimg}.


\section{CONCLUSION}
\vspace{-0.15cm}
In conclusion, we present a novel framework for manipulating deformable and rigid objects using only an RGB camera, eliminating the need for complex sensors, mechanical models, or specialized grippers. Our control strategy adapts the gripper width based on a dissimilarity measure and a scaling factor of tracking points on the object. The dissimilarity correlates with the applied force on deformable objects, while the scaling factor correlates with the object's distance from the camera. Tested on the Franka Emika Research 3 robotic arm and gripper, our system effectively handles various objects, from soft items like lettuce and mozzarella cheese to rigid plastic bottles. The framework's success lies in integrating real-time object detection, segmentation, point tracking, and a grasp controller that adapts to varying levels of object compliance.
\bibliographystyle{IEEEtran}
\bibliography{ref_v}

\end{document}